\documentclass[runningheads]{llncs}
\usepackage{graphicx}
\usepackage{cite} 
\usepackage{times}
\usepackage{epsfig}
\usepackage{amsmath}
\usepackage{amssymb}
\usepackage{mwe}
\usepackage{acro}
\usepackage{amssymb}
\usepackage{xcolor,colortbl}
\usepackage{tabularx}
\usepackage{relsize}
\usepackage{pifont}
\usepackage{booktabs} 
\usepackage{multirow}
\usepackage{multicol}
\usepackage{adjustbox}
\usepackage{float}
\usepackage[colorlinks,
            linkcolor=red,  
            anchorcolor=blue, 
            citecolor=green,   
            ]{hyperref}
\usepackage{amsmath}

\usepackage[utf8]{inputenc}
\begin{document}

\title{Feasibility of Universal Anomaly Detection without Knowing the Abnormality in Medical Images}
\author{Can Cui\inst{1} \and
Yaohong Wang\inst{2} \and
Shunxing Bao\inst{1} \and
Yucheng Tang\inst{3} \and
Ruining Deng\inst{1} \and
Lucas W. Remedios\inst{1} \and
Zuhayr Asad\inst{1} \and
Joseph T. Roland\inst{2} \and
Ken S. Lau\inst{2} \and%
Qi Liu\inst{2} \and
Lori A. Coburn\inst{2} \and
Keith T. Wilson\inst{2} \and
Bennett A. Landman\inst{1} \and
Yuankai Huo\inst{1}}

\institute{
Vanderbilt University, Nashville TN 37235, USA \and
Vanderbilt University Medical Center, Nashville TN 37215, USA 
\and
NVIDIA Corporation, Santa Clara and Bethesda, USA\\
}

\maketitle  

\begin{abstract}
Many anomaly detection approaches, especially deep learning methods, have been recently developed to identify abnormal image morphology by only employing normal images during training. Unfortunately, many prior anomaly detection methods were optimized for a specific ``known" abnormality (e.g., brain tumor, bone fraction, cell types). Moreover, even though only the normal images were used in the training process, the abnormal images were often employed during the validation process (e.g., epoch selection, hyper-parameter tuning), which might leak the supposed ``unknown" abnormality unintentionally. In this study, we investigated these two essential aspects regarding universal anomaly detection in medical images by (1) comparing various anomaly detection methods across four medical datasets, (2) investigating the inevitable but often neglected issues on how to unbiasedly select the optimal anomaly detection model during the validation phase using only normal images, and (3) proposing a simple decision-level ensemble method to leverage the advantage of different kinds of anomaly detection without knowing the abnormality. The results of our experiments indicate that none of the evaluated methods consistently achieved the best performance across all datasets. Our proposed method enhanced the robustness of performance in general (average AUC 0.956).



\end{abstract}

\begin{figure}[t]
\begin{center}
\includegraphics[width=1\linewidth]{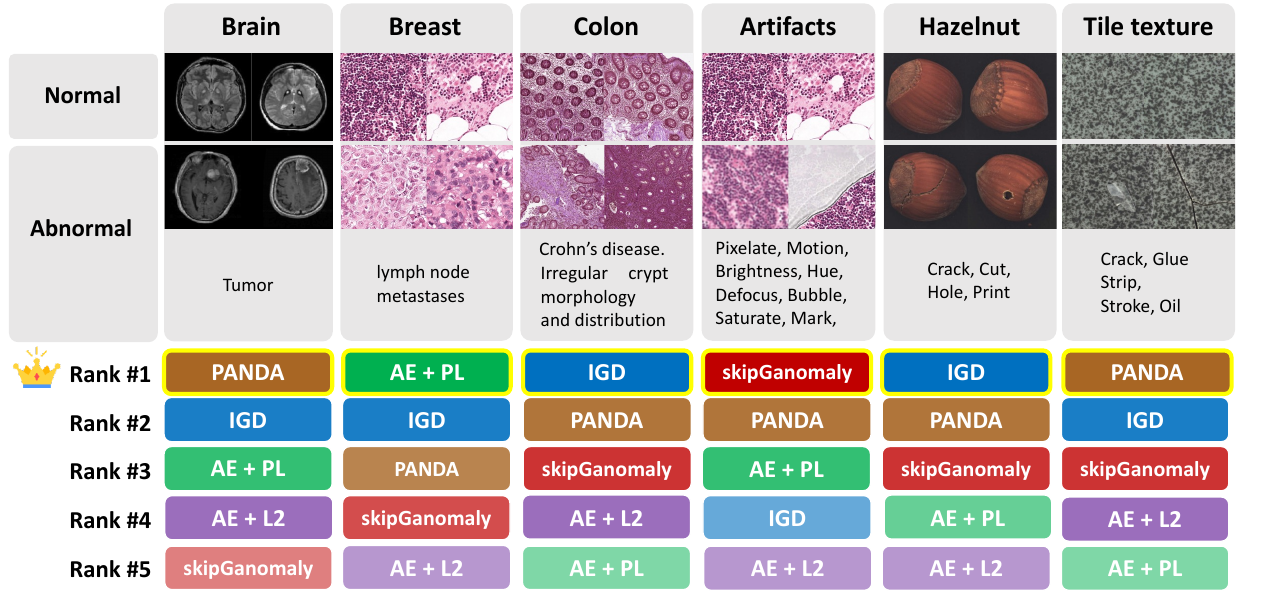}
\end{center}
   \caption{The upper panels show the 4 medical image datasets and 2 natural image datasets that are used for anomaly detection in this study. The lower panels show the ranking of their performance across different image cohorts (corresponding to the results in Table 2). None of these methods consistently achieved the best performance across all datasets.}
\label{fig1:Problem Definition}
\end{figure}

\section{Introduction}
In the context of human perception, it is observed that individuals possess the ability to summarize and store normal patterns, thereby enabling them to recognize the abnormal patterns upon first encounter by comparing them to the normal patterns stored in memory. Especially when existing certain abnormal cases are infrequent or unknown, the ability to discriminate between normal and abnormal patterns must be acquired through learning from normal data. This has driven the research in the area of anomaly detection in machine learning, which has been further enhanced by the advances in deep learning techniques, improving the generalization ability of more complex patterns and leading to more effective detection of abnormality. Different from the regular classification problems, anomaly detection is a kind of one-class classification, where normal and abnormal patterns are binary classifications, but trained solely on normal data and are tested on their ability to detect abnormal patterns ~\cite{yang2021visual, salehi2021unified}.

In the medical domain, vast amounts of data are routinely processed, with the identification of abnormal cases being of great value. For instance, images with poor quality or artifacts should be discarded or require repetition, and patterns in images that deviate from normal patterns may indicate a rare disease. Due to the scarcity of labeled data and the infrequency of abnormal cases, conventional classification methods may not be appropriate. Therefore, there is a need for the development of anomaly detection techniques for medical image data.

Numerous anomaly detection methods have already been proposed, with distribution-based and pretext-task-based strategies being two of the primary approaches ~\cite{yang2021visual}. Distribution-based methods estimate the distribution of normal data or compact the feature space of normal data, allowing abnormal data that lies outside the distribution or boundary of normal data to be recognized as abnormal. Examples of distribution-based methods include One-Class Support Vector Machines (OCSVM) ~\cite{li2003improving}, Deep Support Vector Data Description (DeepSVDD) ~\cite{ruff2018deep}, and Variational Autoencoder (VAE) ~\cite{an2015variational}. On the other hand, pretext-task-based methods involve training models for specific tasks, such as reconstruction ~\cite{gong2019memorizing}, inpainting ~\cite{pirnay2022inpainting}, and denoising ~\cite{kascenas2022denoising}, etc., using normal data only. It is expected that the model can achieve good performance in completing these tasks using normal data but perform poorly when presented with abnormal data. Previous works have explored the effectiveness of pretext-task-based methods for anomaly detection.

According to such prior arts, \textbf{the abnormality in medical image analysis is very heterogeneous and complicated}, including, but not limited to, the existence, shape, density of anomalies and other 
sensory abnormalities. The abnormalities in sensory attributes can also lead to content abnormalities, such as local abnormalities (use pixel-wise reconstruction loss) or more global abnormalities (use perceptual loss at the content level). Unfortunately, \textbf{most of the prior anomaly detection methods were optimized for a specific ``known" abnormality} (e.g., brain tumor, bone fraction, cell types) ~\cite{cai2022dual, shvetsova2021anomaly}. Moreover, even though only the normal images were used in the training set, the abnormal images were often employed during the validation process (e.g., epoch selection, hyper-parameter tuning) ~\cite{shvetsova2021anomaly}, which might leak the supposed ``unknown" abnormality unintentionally. Whether an anomaly detection strategy can perform consistently well for various kinds of anomalies is a problem (Fig.1).  

In this study, we compare the performance of multiple representative anomaly detection methods and introduce a decision-level ensemble to take advantage of different methods to capture multiple kinds of anomalies. Meanwhile, an overlooked but crucial problem in training an anomaly detection model is the hyperparameter selection of the training epoch ~\cite{reiss2021panda}. The selection of a suitable training epoch can influence the results significantly but many works only empirically set a fixed number of epochs ~\cite{akcay2019ganomaly, pmlr-v80-ruff18a, chen2022deep}. Different from the regular classification problems, there may be no anomaly data available in the training phase, so the classification accuracy of the validation set may not be applicable here. Also, the validation set may introduce the bias to known abnormality. In this work, we investigated different epoch selection strategies including a fixed number of epochs, loss of normal data in the validation set, and a dynamic epoch selection method proposed by Reiss et al.~\cite{reiss2021panda}. And their performance was compared with the model selected by a validation set with both normal and abnormal data available.

The contribution of this paper is threefold:

$\bullet$ Firstly, we compare multiple representative anomaly detection methods on various medical datasets. 

$\bullet$ Secondly, we investigate the inevitable but often neglected issues on how to unbiasedly select the optimal anomaly detection model during the validation phase using only normal images.

$\bullet$ Thirdly, we propose a simple decision-level ensemble method to leverage anomaly detection without knowing the abnormality. Extensive experiments were done on 6 datasets and 5 different anomaly detection methods.



\begin{figure*}[t]
\begin{center}
\includegraphics[width=0.7\linewidth]{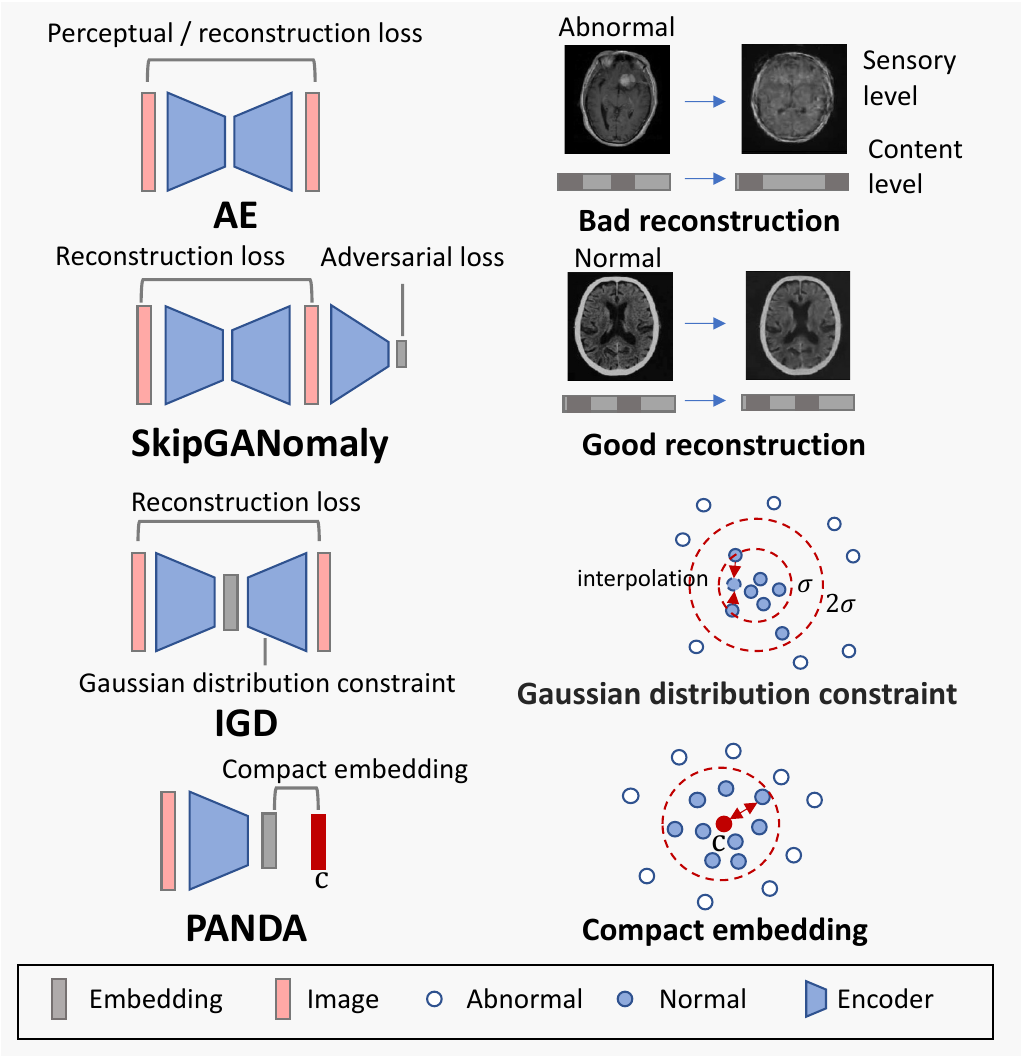}
\end{center}
   \caption{This figure presents the representative anomaly detection methods.}
\label{fig2:Models}
 \end{figure*}

\section{Methods}


\subsection{Anomaly detection benchmarks}
In this work, we selected and compared five representative anomaly detection methods for image data. The structures of these methods are displayed in Fig.2. 

\textbf{1) Autoencoder (AE) with pixel-wise loss.} The autoencoder is a basic reconstruction-based method that aims to train a model to be able to reconstruct normal images but performs worse in reconstructing abnormal images. Images are reconstructed from a bottle-neck feature vector downsampled by convolutional and pooling layers. The pixel-wise loss is used to supervise the reconstruction training and used as the anomaly score.  

\textbf{2) Autoencoder with a perceptual loss.} Instead of using the pixel-wised reconstruction loss. The perceptual loss cares about the content similarity of real and reconstructed images. Especially for pathology images, the dense regions with edges are hard to reconstruct, which may confuse the reconstruction errors results from the anomaly input. The similarity in higher-level embedding spaces from a pre-trained network can also benefit anomaly detection ~\cite{shvetsova2021anomaly}. 

\textbf{3) SkipGanomaly.} ~\cite{akccay2019skip} SkipGanomaly is an enhanced variant of GANomaly ~\cite{akcay2019ganomaly}, which has been developed to generate normal images through the use of both skip connections and adversarial loss. Compared to the auto-encoder-based method, the generated images were found to be less blurry. Zehnder et al ~\cite{zehnder2022multiscale} applied it successfully in anomaly detection in pathology images of breast cancer. 

\textbf{4) IGD.}~\cite{chen2022deep} The IGD method is based on the density estimation of VAE for anomaly detection. It constrains a smooth Gaussian-shape latent space for normal data with adversarially interpolated training samples. Specifically, it not only constrains the latent space with regular VAE design but also forces the model to predict the interpolation coefficient of normal embeddings.

\textbf{5) PANDA.}~\cite{reiss2021panda} The PANDA method draws inspiration from the DeepSVDD method ~\cite{pmlr-v80-ruff18a}. In PANDA, the encoder used for image embedding is replaced with an ImageNet pre-trained encoder, instead of a learned encoder from reconstruction tasks. The ImageNet pre-trained encoder is preferred over a learned encoder because it already has a strong expression ability. The embedding generated by PANDA is then fine-tuned by a distance loss to a fixed embedding center to compact the embedding space. This approach helps to improve the overall efficiency and effectiveness of the anomaly detection model. Also, an elastic regularization inspired by continual learning was added to combat the collapse of compacted space. The distance of an embedding to the fixed center can be used as the anomaly score.    

The above five representative anomaly detection method covers both the pretext-task-based and the distribution-based anomaly detection methods. By covering multiple perspectives, these methods are expected to be effective in detecting various anomalies. As a result, combining them into an ensemble model can increase the overall robustness of the anomaly detection system.

\subsection{Model selection strategies during the validation stage}
An important yet overlooked issue in anomaly detection is the selection of a suitable training epoch for the model. Stopping the training process at different epochs can lead to significantly varied outcomes, but previous research did not adequately address how the epochs were chosen. Unlike typical classification tasks that use both normal and abnormal data in a validation set for epoch selection, the ideal setting of an anomaly detection model should only see normal images, even during the validation stage. Not only because the abnormal cases can be rare to get, but also because the model is expected to be a real ``unbiased" anomaly detection method that deals with unknown abnormalities. Unfortunately, many prior arts employed abnormal images during the validation phase, which leak the known abnormality to the AI model, while some others set a fixed number of epochs for model training which may not fully exploit the model performance at risk. In this paper, we evaluate (1) the performance gap if we only use normal data in both the training and validation phases and (2) how to select the optimal anomaly detection model by only using the normal images.

The first strategy was to employ two sets of normal and abnormal images during the validation stage. Then, select the best model or tuning hyper-parameters based on the best binary classification performance. This was widely used in the prior arts, yet leaked the abnormality to the AI model. Therefore, the models are selected for the best performance on ``known" abnormality.  
Here, we investigate the alternative strategies that only use the normal images during the validation phase:

 \textbf{Strategy} 1) Assessing the loss of normal samples in the validation set, which provides an indication of how well the model has been trained for the pretext task, such as image reconstruction. 

 \textbf{Strategy} 2) Sample-wise early stopping proposed by Reiss et al \cite{reiss2021panda}. Firstly, multiple model checkpoints of different epochs are required to save in the training phase. Then, for each sample in the testing phase, its anomaly score at different checkpoints will be normalized by the corresponding average anomaly score of normal samples in the validation sets and denoted as the maximal ratio. The model checkpoint with a maximal ratio indicates the checkpoint has the best separation to this testing sample and the maximal ratio is used as the anomaly score of this sample.

\subsection{Model Ensemble}
To create an ensemble of different anomaly detection methods, the anomaly scores range of model $i$ on normal data in the validation set ${max(N_{i}) - min(N_{i})}$ is used to normalize the anomaly score of each testing image $\alpha_{i}$ as the equation Eq.~\eqref{eqn:eq1} shown. This is done to eliminate the unnormalized score with a smaller value from being drowned out by the one with much larger values. Once the normalized anomaly scores $\hat{\alpha_{i}}$ are calculated, the average ensemble strategy is used to combine the scores of $k$ different models (Eq.~\eqref{eqn:eq2}). 
\begin{equation}
\label{eqn:eq1}
\hat{\alpha_{i}}  = \frac{\alpha_{i} - min(N_{i})}{max(N_{i}) - min(N_{i})},
\end{equation}
\begin{equation}
\label{eqn:eq2}
\alpha_{ensemble} = \frac{1}{k} \sum_{i}^{k} {\hat{\alpha_{i}}},
\end{equation}
where $N_{i}$ is the anomaly score set of normal data in the validation set for method i.

\begin{table}[]
\centering
\caption{Datasets and the corresponding data splits used in this work}
\begin{tabular}{c|c|cc|cc}
\hline
\multirow{2}{*}{Dataset} & Training & \multicolumn{2}{c|}{Validation}        & \multicolumn{2}{c}{Testing}            \\ \cline{2-6} 
                         & Normal   & \multicolumn{1}{c|}{Normal} & Abnormal & \multicolumn{1}{c|}{Normal} & Abnormal \\ \hline
Brain \cite{msoudnickparvar_2021}                   & 1000     & 400                         & 600      & 600                         & 600      \\
Colon                    & 213      & 71                          & 79       & 80                          & 80       \\
Breast  \cite{litjens20181399}                 & 5462     & 2150                        & 2169     & 4000                        & 817      \\
Artifact \cite{litjens20181399}                & 5462     & 2150                        & 2150     & 4000                        & 4000     \\
Hazelnut \cite{bergmann2019mvtec}                & 391      & 9                           & 12       & 31                          & 58       \\
Tile \cite{bergmann2019mvtec}                    & 201      & 29                          & 19       & 33                          & 65       \\ \hline
\end{tabular}
\label{table:Datasets}
\end{table}

\section{Experiments}
\subsection{Dataset}
Four medical datasets and two natural image datasets are used to investigate and evaluate the anomaly detection algorithm in this work. 1) \textbf{Camelyon breast dataset} ~\cite{litjens20181399}. A public dataset for breast cancer metastase detection in digital pathology. Following the previous work \cite{shvetsova2021anomaly}, patches in the size of 768$\times$768 were tiled from either the healthy tissue or tumor tissue under 40$\times$ magnification and used as normal and abnormal data separately. 2) \textbf{Inhouse colon dataset.} A private pathology dataset of healthy colon tissues and Crohn's disease. The patches tile in the size of 1812$\times$1812 were labeled by pathologists to be normal and abnormal (with disease). 3) \textbf{Camelyon dataset with artifacts.} Nine kinds of common artifacts/corruptions for pathology images were generated on the normal images on the Camelyon dataset ~\cite{litjens20181399} using the toolbox released by Zhang et al. ~\cite{zhang2022benchmarking}. 4) \textbf{Brain tumor dataset}. A public dataset contains 2D MRI slices with tumors or without tumors, in the size of 512$\times$512. ~\cite{msoudnickparvar_2021}  5) \textbf{Hazelnut dataset.} A prevalent computer vision anomaly detection benchmark from MVet ~\cite{bergmann2019mvtec}. 6) \textbf{Tile dataset}. A prevalent computer vision anomaly detection benchmark from MVet ~\cite{bergmann2019mvtec}. 

Image patches in training, validation and testing sets were split by patients. The number of data is shown in Table 1.

\subsection{Experimental Setting}

The experiments were divided into three parts. 

\textbf{1) Comparison of different anomaly detection models.} In section 2.1, we introduced five anomaly detection methods, which were separately applied to 6 datasets. Each method was trained for 250 epochs, with checkpoints saved every 25 epochs, except for the PANDA method, which was trained for 20 epochs with a checkpoint saved every 2 epochs. For the AE-based methods, the architecture used by Cai et al.~\cite{cai2022dual} was employed in this work. The autoencoder with a bottleneck structure consists of 4 down/up convolutional blocks. When the input images were resized to 64 x 64 or 256 x 256, the length of the bottleneck was 16 and 128 separately. The output of the first convolutional layer in the fourth block of the ImageNet pre-trained vgg-16 was used for the perceptual loss to train the reconstruction networks. For the SkipGANanomaly, IGD, and PANDA methods, the official GitHub repositories of these papers were used in this study. The default parameters and experimental configurations outlined in the original papers were adopted unless otherwise specified.

\textbf{2) Comparison of four training epoch selection strategies.}
The most common methods are setting a fixed number of training epochs and selecting the epoch in which the complete validation set (with both normal and abnormal data) achieved the highest performance. They were compared with the sample-wise model selection and strategy using the loss of normal validation mentioned in Section 2.2. 

3) \textbf{Comparison of ensemble model and individual models.} The five individual anomaly detection models were ensembled as Section 2.3 introduced.  

Moreover, all images were processed as a three-channel input, with the channels copied for MRI grayscale images, and then normalized to intensity [0,1]. The batch size for 64 x 64 resolution and 256 x 256 resolution were 64 and 8 separately. To evaluate the capability of the model in discerning normal and abnormal data, the Receiver Operating Characteristic Area Under the Curve (ROC-AUC) score was utilized.

\begin{table}[]
\begin{center}
   \caption{Comparsion of individual and ensembled anomaly detection methods. The AUC scores are computed from the withheld testing data. \textcolor{blue}{\textbf{Blue}} values indicate the best individual model for each cohort. \textcolor{red}{\textbf{Red}} value indicates the best overall performance across all cohorts.(The individual models are selected by the sample-wise early stopping.)} 
\begin{tabular}{l|p{9mm}p{9mm}p{9mm}p{12mm}p{12mm}p{9mm}|c}

\hline
                             Method
                             & Brain & Colon & Breast & Artifact 
                             & Hazelnut & Tile  & Average AUC         \\ \hline
AE + PL (64) ~\cite{shvetsova2021anomaly} & 0.945 & 0.904           & 0.658    & 0.636                 & 0.947    & 0.889 & \textbf{0.830}          \\
AE + L2 (64)          & 0.776 & 0.472           & 0.302    & 0.507                 & 0.860     & 0.719 & \textbf{0.606}          \\
AE + PL (256) ~\cite{shvetsova2021anomaly} & 0.888 & 0.020            & \textcolor{blue}{0.962}    & 0.696                 & 0.930    & 0.670  & \textbf{0.694}          \\
AE + L2 (256)         & 0.711 & 0.224           & 0.108    & 0.506                 & 0.876    & 0.673 & \textbf{0.516}          \\
SkipGanomaly  ~\cite{akccay2019skip}               & 0.660  & 0.634           & 0.815    & \textcolor{blue}{0.999}                 & 0.942    & 0.776 & \textbf{0.804}          \\
IGD ~\cite{chen2022deep}                         & 0.934 & \textcolor{blue}{0.981}           & 0.918    & 0.557                 & \textcolor{blue}{0.965}    & 0.963 & \textbf{0.886}          \\
PANDA ~\cite{reiss2021panda}                        & \textcolor{blue}{0.981} & 0.931           & 0.882    & 0.879                 & 0.959    & \textcolor{blue}{0.980}  & \textbf{0.935}          \\ \hline
Avg. Ensemble (256) (Ours)      & 0.944 & 0.744           & 0.932    & 0.993                 & 0.995    & 0.976 & \textbf{0.931}          \\
Avg. Ensemble (64) (Ours)       & 0.959 & 0.931          & 0.867    & 0.995                 & 0.998    & 0.986 & \textcolor{red}{\textbf{0.956}} \\ \hline
\end{tabular}
\end{center}
\text{*AE: auto-encoder,  *PL: perceptual loss, *L2: L2 loss }\\
\text{*(64): 64$\times$64 resolution, *(256): 256$\times$256 resolution}\\
\label{table:Results1}
\end{table}

\begin{table}[]
\begin{center}
\caption{Comparsion of different model selection methods. The values \textbf{in bold} indicate the best performance among the four methods. The \underline{underscore} values indicate the second highest performance.}
\begin{tabular}{l|c|c|c|c}
\hline
                Method
                &             & \multicolumn{2}{c|}{Validation using Only Normal Images}           & Normal \& Abnormal   \\ \cline{2-5} 
                & Last epoch  & Normal Validation & Sample-wise Model & Complete Validation \\ \hline
AE   + PL (64)  & 0.828       & \underline {0.841}       & 0.830             & \textbf{0.848}      \\
AE   + L2 (64)  & 0.547       & 0.562             & \underline{0.606}       & \textbf{0.644}      \\
AE   + PL (256) & 0.655       & \textbf{0.719}    & 0.694             & \underline{0.718}         \\
AE   + L2 (256) & 0.446       & 0.471             & \underline{0.516}       & \textbf{0.583}      \\
SkipGanomaly    & 0.648       & 0.669             & \underline{0.804}       & \textbf{0.875}      \\
IGD             & 0.837       & 0.855             & \textbf{0.886}    & \textbf{0.886}      \\
PANDA           & \underline{0.942} & \underline{0.942}       & 0.935             & \textbf{0.944}     \\

\hline
\end{tabular}
\end{center}
\label{table: Results2}
\end{table}

\section{Results and Discussion}
The results of the ensemble anomaly detection are compared with the individual methods and presented in Table 2. It can be seen that none of the individual anomaly detection methods outperforms other methods across all datasets. Even though some methods achieved the best performance in some datasets, they may fail in some other datasets. The average ensemble results take advantage of anomaly detection from different aspects and tend to achieve more robust results. Notably, the ensemble method surpasses the best models in Hazelnut and Tile datasets and demonstrates competitive performance in other datasets. Moreover, the average AUC of the ensembled method across all datasets can outperform the best individual anomaly detection method in our experiments.
Table 3 is the comparison of different model selection methods for the best training epochs. It is observed that with the help of labeled abnormal data in the validation set, the results always outperformed other methods across most of the datasets. The sample-wise selection method was performed close to the complete validation set, but it requires larger saving memory and multiple times of inferences for better performance. The epoch selected by the validation set with normal data only performed better than the baseline method using the fixed number of epochs, which is an efficient and practical method for model selection in anomaly detection.





\section{Conclusion}
In this paper, we assess the prevalent anomaly detection approaches on six image cohorts. 
Based on the experiment results, we draw the following conclusions: (1) None of the evaluated methods consistently attain the best performance across all datasets. (2) Current model selection methods commonly involve abnormal images during the validation stage, inadvertently disclosing the abnormality and consequently yielding better performance compared to a more stringent model selection approach that uses only normal images during validation. (3) Our proposed simple ensemble method improves anomaly detection performance without requiring knowledge of the abnormality.

\section{Acknowledgements}
This work is supported by the Leona M. and Harry B. Helmsley Charitable Trust grant G-1903-03793, NSF CAREER 1452485.


%
\bibliographystyle{splncs04}
\bibliography{main}
\end{document}